\documentclass[conference]{IEEEtran}
\usepackage{times}

\usepackage[numbers]{natbib}
\usepackage{amsmath}
\usepackage{amsfonts}
\usepackage{amssymb}
\usepackage{graphicx}
\usepackage{multicol}
\usepackage{multirow}
\usepackage[bookmarks=true, colorlinks=true]{hyperref}
\hypersetup{urlcolor=blue, citecolor=blue, linkcolor=blue}
\usepackage{graphicx}
\usepackage{subcaption}

\begin{document}

\title{EDEN: Entorhinal Driven Egocentric Navigation Toward Robotic Deployment}

\author{
    \authorblockN{
        Mikolaj Walczak\authorrefmark{1}, 
        Romina Aalishah\authorrefmark{1}, 
        Wyatt Mackey\authorrefmark{2},
        Brittany Story\authorrefmark{2},
        David L. Boothe Jr.,\authorrefmark{2},\\
        Nicholas Waytowich\authorrefmark{2},  
        Xiaomin Lin\authorrefmark{1},
        Tinoosh Mohsenin\authorrefmark{1}
    }
    \authorblockA{\authorrefmark{1}Johns Hopkins Whiting School of Engineering, Baltimore, Maryland, United States\\
    Emails: \{mwalcza1, raalish1, xlin52, tinoosh\}@jhu.edu}
    \authorblockA{\authorrefmark{2}DEVCOM United States Army Research Laboratory, Aberdeen, Maryland, United States\\
    Emails: \{wyatt.t.mackey, brittany.m.story, david.l.boothe7, nicholas.r.waytowich\}.civ@army.mil}
}

\newcommand{\sys}{EDEN}


%

\maketitle

\IEEEpeerreviewmaketitle

\begin{abstract}
Deep reinforcement learning agents are often fragile while humans remain adaptive and flexible to varying scenarios. To bridge this gap, we present \sys{}, a biologically inspired navigation framework that integrates learned
entorhinal-like grid cell representations and reinforcement
learning to enable autonomous navigation. Inspired by the mammalian entorhinal-hippocampal system, \sys{} allows agents to perform path integration and vector-based navigation using visual and motion sensor data.
At the core of \sys{} is a grid cell encoder that transforms egocentric motion into periodic spatial
codes, producing low-dimensional, interpretable embeddings of position. To generate these
activations from raw sensory input, we combine fiducial marker detections in the lightweight MiniWorld simulator and DINO-based visual features in the high-fidelity Gazebo simulator.
These spatial representations serve as input to a policy trained with Proximal Policy
Optimization (PPO), enabling dynamic, goal-directed navigation.
We evaluate \sys{} in both MiniWorld, for rapid prototyping, and Gazebo, which offers realistic
physics and perception noise. Compared to baseline agents using raw state inputs (e.g., position, velocity) or standard convolutional image encoders, \sys{} achieves a 99\% success rate, within the simple scenarios, and \textgreater 94\% within complex floorplans with occluded paths with more efficient and reliable step-wise navigation. In addition, as a replacement of ground truth activations, we present a trainable Grid Cell encoder enabling the development of periodic grid-like patterns from vision and motion sensor data, emulating the development of such patterns within biological mammals. This work represents a step toward biologically grounded spatial intelligence in robotics, bridging neural navigation principles with reinforcement learning for scalable deployment. A publicly available GitHub repository for \sys{} is made available at \href{https://github.com/M-iki/EDEN/tree/main}{\texttt{github.com/M-iki/EDEN}}.

\end{abstract}
\section{Introduction}

Accurate localization and efficient path planning are fundamental components of any autonomous navigation system. Localization determines the position and orientation of the agent relative to a global reference frame, while path planning generates collision-free trajectories toward target positions. Without reliable estimates of position and direction, a robot cannot avoid obstacles, reach its target, and adapt when the scenario changes. Small errors in estimating position or direction can accumulate over time and lead to task failure. 

\begin{figure}
    \centering
    \includegraphics[width=0.63\linewidth]{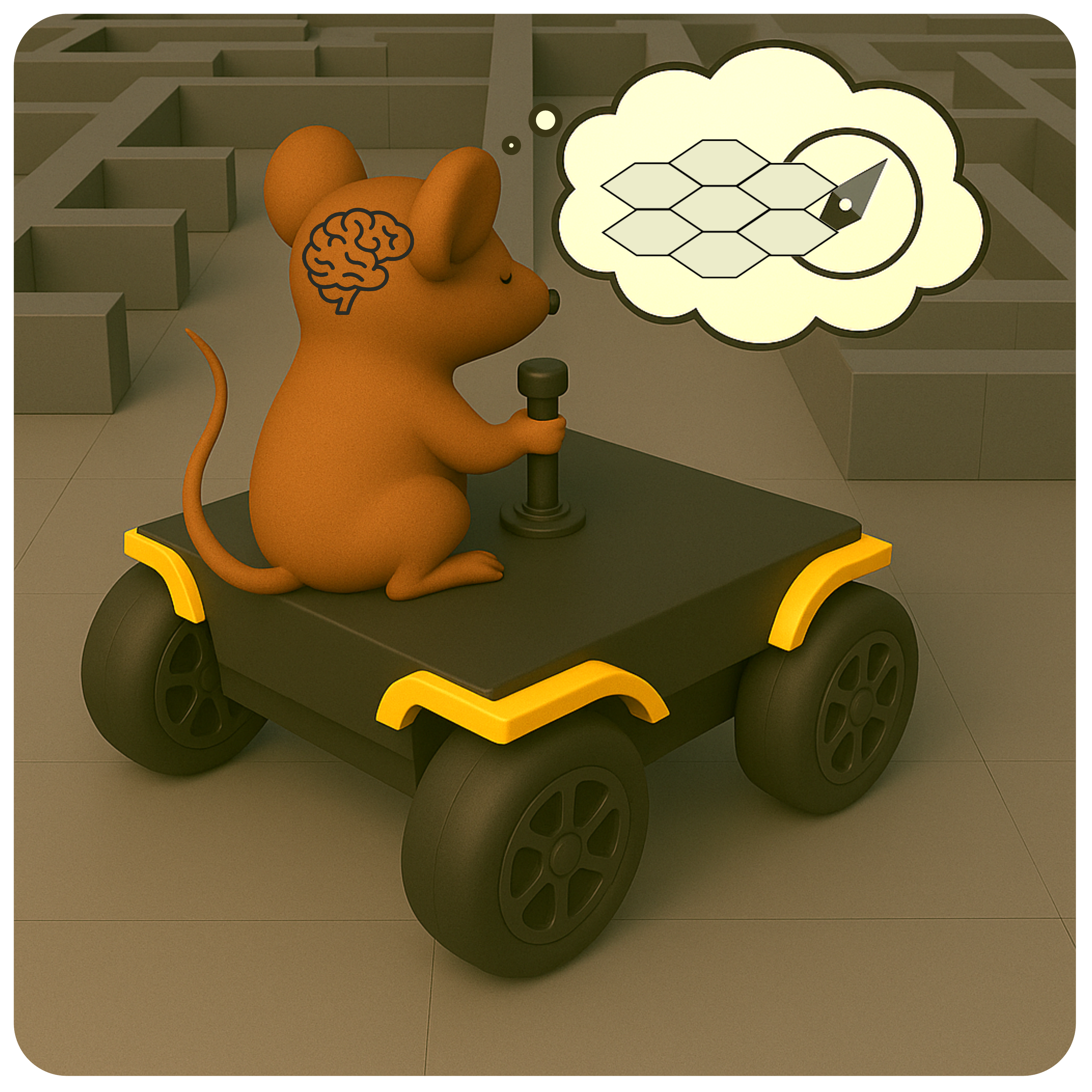}
    \caption{Representation of EDEN inspiration. The proposed navigation system is inspired by grid cells found in the entorhinal cortex of the mammalian brain, which support spatial reasoning and goal-directed behavior. \textit{The conceptual visualization was generated with the assistance of OpenAI’s DALL·E.}}
    \label{fig:intro_figl}
\end{figure}

To address these challenges, most robotic systems rely on Simultaneous Localization and Mapping (SLAM) or visual odometry, which estimate the position of the agent by building a map of the scenario and updating it using sensor inputs such as LiDAR, depth, or camera frames. Once the map is available, motion planners compute collision-free paths within it. Reinforcement learning (RL) has emerged as an alternative approach, allowing agents to learn navigation behaviors through direct interaction with their surroundings~\cite{kober2013reinforcement, Singh2022Reinforcement, Tang2025DeepRL}. RL agents learn action policies directly from observations without requiring an explicit map, which makes them more flexible in dynamic or partially-observable settings.

However, both of these approaches suffer from significant drawbacks. Several studies have shown that SLAM and visual odometry suffer from temporal drift, especially in scenarios with repetitive sensor data, eg.~images~\cite{cadena2016past, mur2015orb}. This error accumulates over time, leading to localization drift and degraded performance in planning tasks. Conversely, purely RL-based navigation policies often lack structure and poorly generalize, as they depend entirely on end-to-end learning and do not maintain a consistent representation of space. This results in inefficiencies and difficulties in long-horizon planning.

In contrast, animals navigate using flexible, extensible learned spatial representations. Hafting et al.~\cite{hafting2005microstructure} found that the dorsocaudal medial entorhinal cortex (dMEC) in mammalian brains creates an internal cognitive map using special neurons called grid cells. These cells encode space through a multi-scale periodic representation that supports updating one’s position using self-motion cues (path integration) and computing direct routes to goals (vector-based navigation) by comparing grid representations. Built on this, several studies have explored computational models inspired by grid-cell mechanisms~\cite{cueva2018emergence, sorscher2023unified, liao2024cognitivemap, burak2009accurate, stemmler2015connecting}. For example, Banino et al.~\cite{banino2018vector} trained a recurrent network to perform path integration, leading to the emergence of grid cell-like representations. This provided a foundation for efficient, vector-based navigation in complex scenarios. These also suggest that brain-inspired representations could enhance robotic navigation when integrated with learning frameworks. Recent work has further extended this approach to robotics, demonstrating that brain-inspired architectures using deep learning-emerged grid cell models can be deployed for navigation in both simulated and physical scenarios~\cite{wang2023bioinspired, ivanovs2025deep, yu2019bionic}.

Unlike SLAM, which builds and stores a detailed map using LiDAR and point clouds, humans learn to navigate by tracking landmarks and memorizing object locations. London taxi drivers have even shown structural growth in their hippocampus as they develop a full mental map of the city~\cite{maguire2000navigation}. Similarly, \sys{} uses brain-inspired grid representations to keep track of its position internally and uses landmarks to reach the goal, storing important high level representations of space rather than estimated point clouds. 


\begin{figure*}[t]
    \centering
    \includegraphics[width=0.85\textwidth]{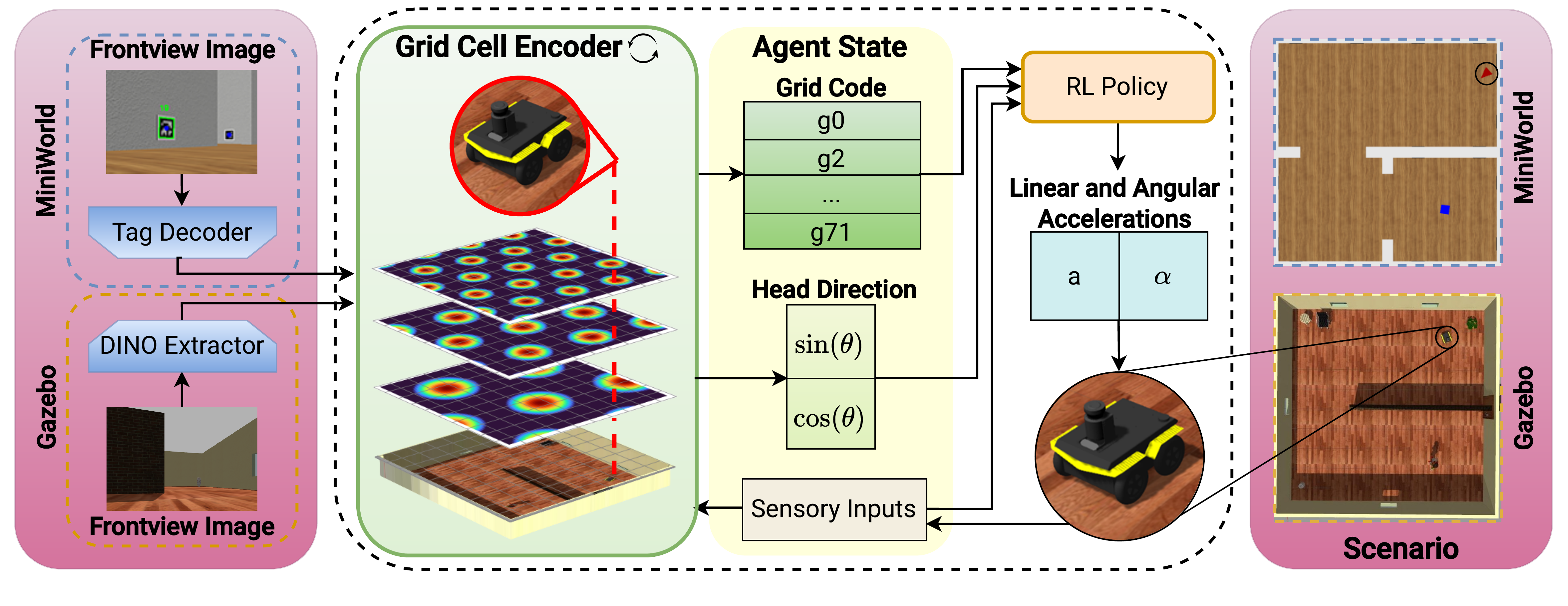}
    \caption{Proposed architecture of \sys{} deployed within the MiniWorld and Gazebo simulators. Grid cell encodings are developed from features extracted using front-facing views and velocity signals, features created in MiniWorld via detected fiducial tags, replaced by general-purpose features extracted using the DINO framework within Gazebo. The predicted grid cell activations and head direction are then combined with sensory inputs passed to the RL policy for controlling the agent via acceleration commands.}
    \label{fig:architecture}
\end{figure*}

Motivated by these biological insights, we propose a brain-inspired navigation framework that combines the strengths of both neuroscience and machine learning as shown in Figure~\ref{fig:intro_figl}. Instead of relying on external maps or raw policies, \sys{} embeds an internal grid-cell-like representation that is updated by velocity and visual landmarks. This representation serves as input to an RL policy for decision-making. As a result, the agent maintains an internal sense of position and learns efficient navigation strategies through rewards. Note this is different from traditional sinusoidal position encoding schemes, since the position encodings are inferred from self motion and visual cues.

Our contributions are summarized as follows:

\begin{itemize}

    \item We design and train a an RL policy architecture that uses grid representations inspired by the mammalian entorhinal cortex. Our architecture  enables efficient navigation under noisy or partial observations without relying on pre-defined external maps.

    \item We develop a grid cell encoder network distilled from ground truth activations, which enables the prediction of grid cell responses from high-level visual features and proprioceptive signals.
    
    \item We evaluate \sys{} in both simple and complex scenarios using lightweight and high-fidelity simulators. We compare \sys{}'s performance against baselines that rely on explicit position data, raw visual inputs, and preprocessed object detection.
    
\end{itemize}

\section{Related Work}
\subsection{Simultaneous Localization and Mapping (SLAM)}
SLAM is a longstanding standard for flexible autonomous navigation~\cite{wang2024slam, Chen2025}. It uses data from multiple sensors, such as LiDAR, IMU, GPS, and cameras, which are merged to help mitigate the weaknesses of any single modality~\cite{yin2022multi}. Shin et al. introduced a direct-vision SLAM method, showing that integrating LiDAR and vision camera data can enhance accuracy and real-time performance when sparse depth information is combined~\cite{Shin2018}. Herraez et al. achieved smooth and accurate onboard simultaneous localization and mapping by combining automotive radars with an IMU and exploiting the additional velocity and radar cross-section information provided by radar sensors~\cite{CasadoHerraez2025}.
However, despite the powerful aspects of SLAM, approaches in this area can suffer from temporal drift, especially in visually repetitive scenarios~\cite{cadena2016past, mur2015orb}. Errors accumulate over time, leading to localization drift and degraded task planning performance.  Furthermore, components used in SLAM can be noisy, and introduce inaccuracies to the calculations~\cite{yan2024fault}. Even though there exist several denoising filters, such as the Kalman filter~\cite{kalman1960new}, they have have low flexibility and adaptability under complex conditions~\cite{barshalom2001estimation}. 

\subsection{Path planning}
Path planning is another key component of effective autonomous navigation~\cite{Gasparetto2015, Zhang2018}. Classical algorithms such as A*~\cite{astar} and D*~\cite{dstar} offer reliable solutions in well-mapped scenarios~\cite{Liu2022, Ju2020} but often require full observability and accurate localization, which may not be possible in unknown scenarios. Song et al. developed a model that learns human driving behavior and adapts to different roads by a combination of CNN and LSTM-based networks~\cite{Song2018}. More recent approaches based on deep reinforcement learning (DRL) have shown promise in learning end-to-end policies that combine perception, localization, and control~\cite{mirowski2016learning, alharthi2025novel}. These methods enable agents to navigate complex scenarios without access to maps, instead using learned spatial representations and raw sensor inputs. However, without reliable cues, DRL agents often struggle with drift and complexities~\cite{miranda2023generalization}. Benchmarks such as Habitat point out that many policies that excel in their training mazes collapse when the geometry or texture distribution shifts~\cite{savva2019habitat}.

\subsection{Bio-inspired navigation architectures}
Recent research has begun to explore bio-inspired solutions, particularly mammalian grid cells, which support spatial and stable localization in biological systems. Notably, \emph{in vivo} grid cells can maintain structured representations even in darkness for up to \~10 minutes~\cite{hafting2005microstructure}. This highlights their role in consistent localization across time and sensory uncertainty. Grid cell-based models have been shown to enable agents to form internal maps and localize themselves effectively~\cite{banino2018vector, tang2018cognitive}. Yang et al.~showed a possible neural solution to overcome the serious drift of IMU-based inertial navigation of Unmanned Aerial Vehicles (UAVs) in the absence of external sensory cues~\cite{burak2009accurate}. Later, Sheng et al.~extended the role of the grid cell system from spatial navigation to visual concept space, meaning the information retrieved can be used for situational awareness~\cite{sheng2021hippo}. 


\section{Proposed Approach}
This section outlines the approach used to implement \sys{} within two different simulators: MiniWorld~\cite{MinigridMiniWorld23} and Gazebo~\cite{koenig2004design}, as illustrated in Figure~\ref{fig:architecture}. 

\subsection{Grid Cell and Head Direction Encoder}
The grid and head direction modules maintain a continuous estimate of the position and orientation of the agent through a combination of egocentric motion cues, such as linear and angular velocities, and visual features to update a latent spatial code over time.

The core of the encoder is an LSTM operating on the MLP-transformed motion and sensory features. The output is a compact embedding that encodes the location of the agent in a distributed representation. This embedding is concatenated with other sensory inputs and passed to the policy network to produce the corresponding action. The encoder undergoes supervised pretraining using idealized grid representatives
\begin{align*}
    \alpha \max\Bigg(0,\; 
        &\cos(\hat{x} - \phi_x) 
        + \cos\left( \frac{-\hat{x} + \phi_x}{2} + \frac{(\hat{y} + \phi_y)\sqrt{3}}{2} \right) \\
        &+ \cos\left( \frac{-\hat{x} + \phi_x}{2} - \frac{(\hat{y} - \phi_y)\sqrt{3}}{2} \right)
    \Bigg),
\end{align*}
\noindent where $\phi$ is a phase shift, $\alpha$ is a scaling factor, and $(\hat{x}, \hat{y}) = R(x, y)$ is a rotation of the standard basis. This approach minimizes the gap between the predicted and actual positions over time and enables the encoder to learn a stable mapping from motion sequences to spatial embeddings. Multiple grid code embeddings are generated on various spatial scales $\alpha$, each combined with several phase changes $\phi$ and repeated under different rotational frames $(\hat{x},\hat{y})$. This results in a high-dimensional representation that enables accurate localization through the overlapping of periodic patterns detailed in the Grid Cell Encoder block of Figure~\ref{fig:architecture} with an overview of the implemented module outlined in \ref{subsec:encoder_training}.

Alongside the grid code embedding of position, a periodic encoding is applied to represent the agent’s orientation, $\theta$, relative to the global reference frame. To ensure smooth transitions across angular boundaries (e.g. near $0$ and $2\pi$), the raw angle $\theta$ is transformed as $[\sin(\theta),\cos(\theta)]$. This continuous embedding facilitates stable learning akin to the periodic representation found within head direction cells of biological mammals.

The proposed encoder is fully differentiable and trained end-to-end within the RL pipeline, making it robust to input noise. Importantly, it serves as an alternative to raw coordinate inputs, enabling the agent to reason about space without direct access to positional information, thereby promoting more biologically plausible behavior.

\begin{figure}[h!]
    \centering
    \includegraphics[width=0.95\linewidth]{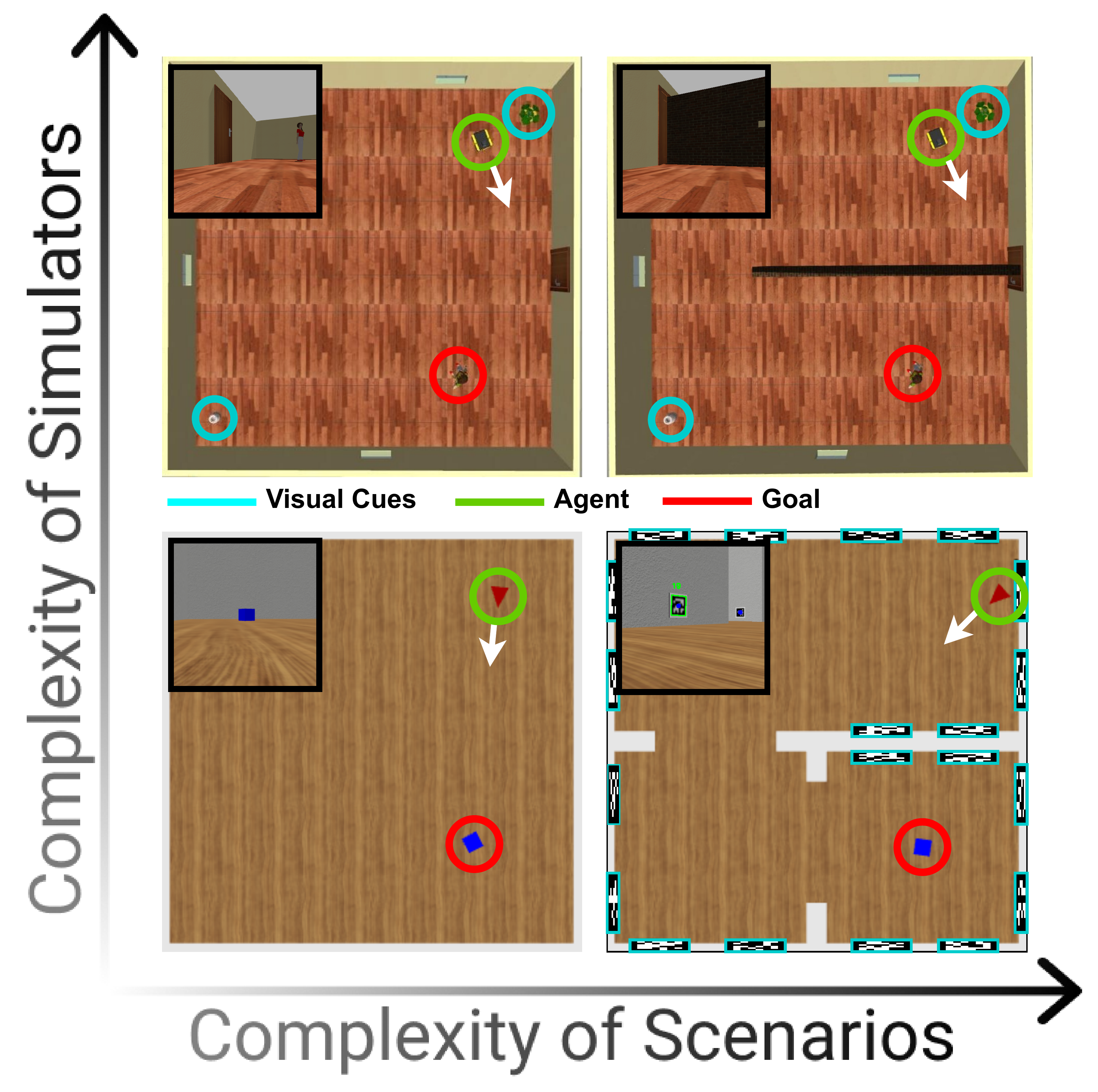}
    \caption{MiniWorld and Gazebo scenarios used for training with both simple and complex floorplans. In MiniWorld, landmarks are defined using unique fiducial tags placed on walls (bottom right), while in Gazebo, these tags are replaced with realistic objects (top left).}
    \label{fig:envs}
\end{figure}

\subsection{Reinforcement Learning Policy and Algorithm}
With the positional encoding of \sys{} established, we adopt the Proximal Policy Optimization (PPO) algorithm~\cite{schulman2017proximal}, implemented via Stable-Baselines3~\cite{stable-baselines3}, as the RL backbone due to its robustness and reliable training performance. We apply a custom reward function that penalizes collisions with the scenario and the use of longer trajectories, while rewarding the agent for reaching the goal.
%


\section{Experimental Setup}
In this section, we outline the experimental setup used to evaluate the proposed system for comparison with baseline input modalities. The section begins with defining task scenarios tailored to two simulators.

\subsection{Simulator and Agent Setup}
Using a stepwise approach, we first deploy and develop \sys{} in the MiniWorld~\cite{MinigridMiniWorld23} simulator—a lightweight platform designed for vision-based navigation tasks. Following this, we transition to the more realistic Gazebo~\cite{koenig2004design} simulator, which offers high-fidelity robotic simulations making it better suited for evaluating the real-world applicability of \sys{}.

Within these simulators, we construct two distinct scenarios to evaluate \sys{}'s performance. In the first, the agent must navigate within an open room. In the second, a wall separates the agent from the goal, requiring planing around occluded paths. In both cases, the agent is initialized at a random starting position and tasked with reaching a fixed goal located at the center of the lower right quadrant. These scenarios are visualized in Figure~\ref{fig:envs}.

While the simulators vary in complexity, both pose challenges related to maintaining long-term spatial consistency and planning under partial observability. In biological systems, such challenges are met by integrating multiple sensory modalities. For instance, in the entorhinal cortex, visual cues are combined with vestibular input to stabilize internal spatial representations over time~\cite{keinath2018environmental}. Inspired by this mechanism, we design agents and scenarios that incorporate visual cues aligned with the agent's motion, allowing \sys{} to maintain a coherent internal estimate of its position and orientation.

\subsubsection{Action Space}
To emulate vestibular input observed in mammals, the agent interacts with the simulation through a continuous action space defined by linear acceleration ($a$) and angular acceleration ($\alpha$) per step. This setup enables both translational and rotational motion, allowing for fine-grained control over the agent’s trajectory and heading. The resulting velocities are then clipped to predefined maximums, normalized, and fed into \sys{}. A description of how these actions are translated into motion commands is outlined in Section~\ref{subsec:abstract_movement_scheme}.

\subsubsection{MiniWorld Agent and Landmarks}
In MiniWorld, the agent is represented as a red triangle. During each step, the agent receives a forward-facing RGB image, which serves as the input for computing Grid Cell activations. To incorporate visual cues, AprilTag fiducial markers~\cite{olson2011apriltag} are placed on the surrounding walls. The tag closest to the center of the agent’s view is identified and its data is used as an additional input.

\subsubsection{Gazebo Agent and Landmarks}
The transition to Gazebo enables training and evaluation of \sys{} in a more realistic and physically grounded simulator. Scenarios are developed using the CPR Gazebo framework~\cite{clearpathrobotics_cpr_gazebo}, which models a Clearpath Jackal robotic agent equipped with a LiDAR sensor and a front-facing RGB camera. Leveraging this framework, along with assets from the Gazebo Models and Worlds collection\cite{yao_gazebo_models_worlds_2025}, an office-like floorplan scenario is constructed.

To enhance realism and promote generalization beyond synthetic visual markers, the AprilTag fiducials used in MiniWorld are replaced with common real-world objects, such as an office chair, a potted plant, and a human figure. Visual features from these objects are extracted using DINOv2~\cite{oquab2023dinov2}, a self-supervised vision transformer model employed as a general-purpose feature extractor. 
 
\begin{figure*}[t]
    \centering
    \includegraphics[width=0.8\textwidth]{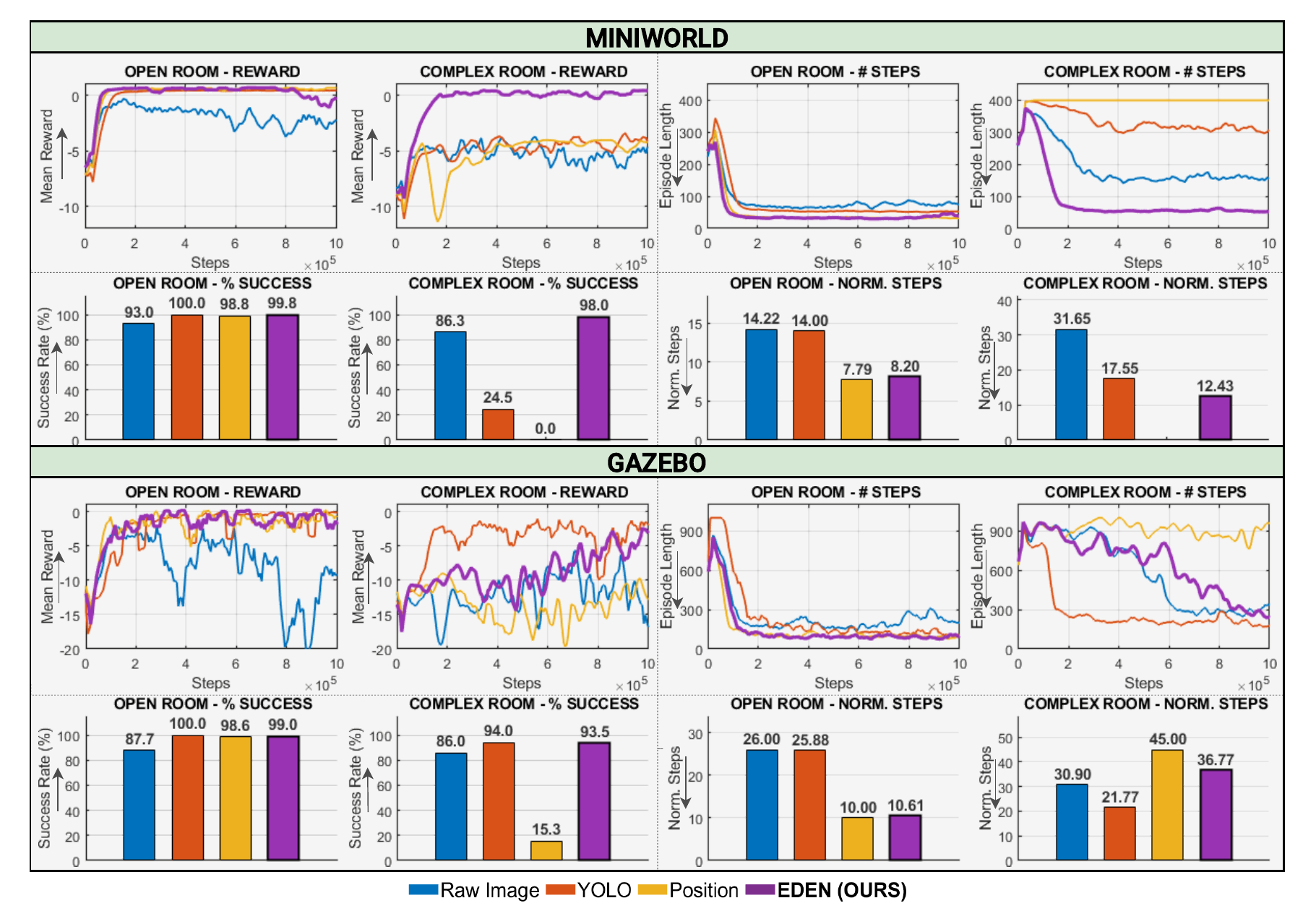}
    \caption{Performance comparison across four training scenarios using three baseline observation modalities and \sys{}. Top: Mean episode reward. Bottom: Success rate and steps normalized by shortest path distance, evaluated on successful episodes within 1000 testing episodes. All modalities receive the agent's current velocity with the addition of LiDAR within Gazebo. Due to there being no successful episodes within the MiniWorld complex scenario given raw position input, the normalized step metric is omitted.}
    \label{fig:miniworld_gazebo_successrate}
\end{figure*}

\begin{figure*}[t]
    \centering
    \includegraphics[width=0.78\textwidth]{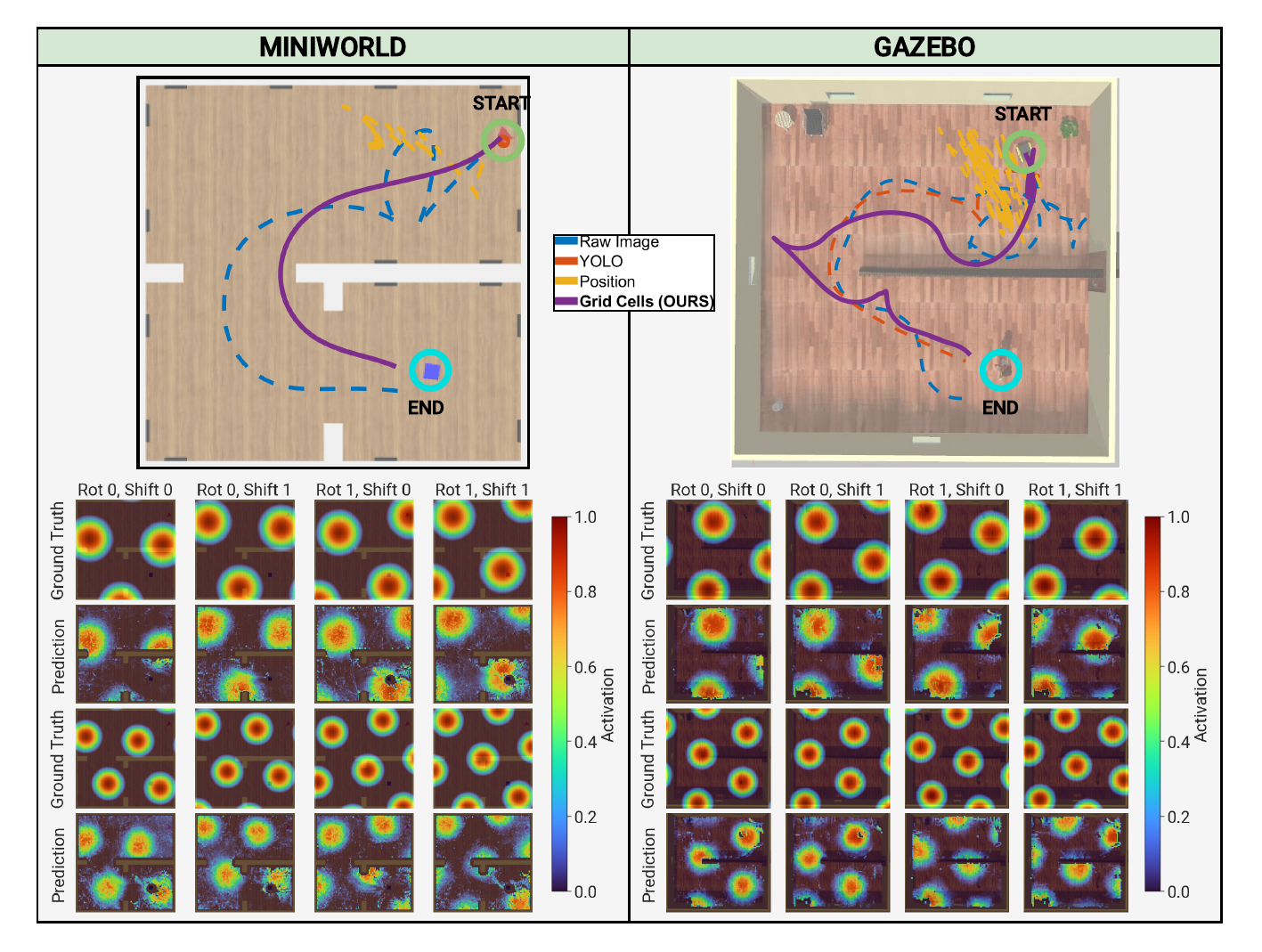}
    \caption{Sample episodes from each of the four trained models performing goal-directed navigation in MiniWorld and Gazebo (top). For each episode, the ground-truth grid cell activations are shown, alongside the predictions obtained by averaging activations from both policy-driven and random trajectories. This comparison illustrates the decoder's ability to reconstruct spatial representations from visual inputs and effectively navigate within the complex scenario.}
    \label{fig:decoder_results}
\end{figure*}

\subsection{Grid Cell Encoder Training}\label{subsec:encoder_training}
We adopt a sequential approach to develop the trainable Grid Cell encoder. First, an initial RL policy is trained using ground-truth grid activations. Once this policy has sufficiently converged, the agent is deployed in repeated randomized episodes to collect vision and motion data. To enhance the encoders reliability, we create an additional dataset using a randomly acting agent within the scenario. Using the full dataset, we train an LSTM network to approximate the grid cell representations, with the training procedure detailed in Section~\ref{subsec:abstract_gc_encoder}. This network, which receives both proprioceptive and visual inputs, serves as a drop-in replacement for the ground-truth grid activations.

\subsection{Baseline Input Modalities}\label{subsec:baselines}
To evaluate the effectiveness of grid cell encoding as the primary localization signal, we introduce a set of baseline input modalities. Each modality receives the normalized linear velocity ($v$) and angular velocity ($\omega$) as proprioceptive input. In Gazebo, we additionally augment all networks with LiDAR point cloud data. The evaluated input modalities are as follows:

\begin{itemize}
    \item \textbf{Position and Orientation.} The raw $[x, y]$ position of the agent, normalized by the scenario's maximum bounds, is provided as input alongside a sine and cosine encoding of the agent's orientation.
    \item \textbf{Raw Image.} The front view image of the agent is extracted and fed directly into the policy.
    \item \textbf{YOLO Detection.} The raw image is passed through the YOLOv5 object detector~\cite{glenn_jocher_2020_4154370}, from which only the detection corresponding to the goal class is provided. 
\end{itemize}
For fair comparison, identical policy architectures are used across all modalities, with the addition of convolutional layers in the raw image baseline for spatial feature extraction. An in detail overview of the policies is provided in Section~\ref{subsec:abstract_reinforcement_policies}.

\section{Results}
We evaluate \sys{} in two key areas: (1) the effectiveness of grid cell activations in training RL policies using cumulative reward, success rates, number of steps, and distance-normalized steps, with detailed calculations outlined in Section~\ref{subsec:abstract_norm_steps}, and (2) the feasibility of encoding visual and motion sensor data into grid cell representations.

\subsection{Reinforcement Learning Agent Performance}
We show success rates and episodic reward for open and complex rooms in both MiniWorld and Gazebo in Figure \ref{fig:miniworld_gazebo_successrate}, including both the grid inspired and baseline models discussed in Section \ref{subsec:baselines}. We additionally show sample paths taken by the different agents in Figure \ref{fig:decoder_results}. Our results show that across all models, the open room is easily and near-instantly resolved (bottom left bar charts of Figure \ref{fig:miniworld_gazebo_successrate}). The exception to this is training with raw pixels, which results in the model learning artifacts that destabilize training. The complex scenario is substantially more difficult, however: when training on limited epochs, ground truth-normalized $(x, y)$ coordinates failed entirely to adapt to the position of a wall. In the MiniWorld set-up, grid activations with learned grid cells are able to outperform both unprocessed and processed images in the complex scenario as well. One interesting point is that, while normalized position has the lowest success rate, it has a comparable reward to raw pixels and YOLO, suggesting that it is superior at avoiding obstacles, at the expense of failing to learn a viable path to the goal. In the Gazebo scenario, in which all models additionally had access to LiDAR data, YOLO-based image preprocessing resulted in the swiftest and most stable training, though grid cell encoding achieved a comparable success rate at one million training steps. 

In the top of Figure~\ref{fig:decoder_results}, we show sample trajectories of the best version of each method in the complex scenarios. In MiniWorld, the YOLO based model never learns a strategy beyond swiveling, and therefore does not move from its position. In Gazebo, however, where YOLO is augmented with a recurrent policy and LiDAR data, it is able to smoothly converge. The raw normalized position fails in complex scenarios, while raw pixels and the grid cell method both ultimately succeed in both scenarios. Grid encodings of our model in Gazebo contain several notable cusps, which we believe are the result of the model not fully converging by 1 million steps. We anticipate that mild additional training would continue to optimize both the raw pixel and grid cell paths. 

\subsection{Developed Grid Cell Encodings}
For final deployment, the ground truth activations used to train the \sys{} RL policies are not directly accessible. To address this, using the procedure outlined in \ref{subsec:encoder_training} we trained an LSTM based grid cell encoder. The bottom half of Figure~\ref{fig:decoder_results} presents a heatmap comparison between the predicted and real grid cell activations across eight representative cells sampled over multiple agent trajectories. The learned activations closely match the ground truth in terms of translation, scale, and rotation, demonstrating that the grid cell encoder effectively captures spatial representations.
\section{Conclusion} 
\label{sec:conclusion}
In this work, we present \sys{} a biologically inspired framework that integrates a localization model and an RL agent inspired by grid cell structures found within the entorhinal cortex. By processing egocentric inputs (specifically velocity signals and visual cues) \sys{} demonstrates effective navigation in both simple and complex scenarios in the MiniWorld and Gazebo simulators. The agent achieves an overall success rate exceeding 94\% across all scenarios, exhibiting both efficient and reliable navigation behavior. Using the trained policies, we further develop a grid cell encoder module capable of accurately predicting ground truth activations solely from image and motion data. 

\sys{} presents a step toward the development of more autonomous and resilient robotic agents by leveraging principles from mammalian neural navigation systems. Through four test scenarios, we evaluate the feasibility of this approach in controlled environments. However, real-world deployment introduces significant complexity and variability, introducing the need for the development of policies that can generalize to dynamic, unfamiliar conditions. Building on this foundation, future work will explore the integration of local coordinate frames and related techniques to support zero-shot generalization in novel scenarios, eliminating the need for predefined maps or recurring spatial structures.
\section*{Acknowledgments}
This work has been partially supported by the Army Research Laboratory Cooperative Agreement No W911NF2120211.


\bibliographystyle{plainnat}
\bibliography{references}
\clearpage
\section{Appendix}
\subsection{Grid Cell Module}\label{subsec:encoder_training}
Within this work we utilize the grid cell encoding equation using a total of 72 differing grid cell activations. These activations are limited by a maximum scale factor $\alpha$ of 10, within Gazebo and 5 within MiniWorld with 3 initial cells. This scale factor is multiplied over 12 shifts which are then further mmultiplied over 2 orientations. Effectively this configuration of 72 relates to the number of cells used within \sys{}. In this work these values were selected as an initial base case which can in the future be replaced with optimal configurations determined based on the current application and complexity of the scenarios.

\subsection{Movement Scheme}\label{subsec:abstract_movement_scheme}

The actions produced by the reinforcement learning policy correspond to an acceleration normalized within $[-1,1]$. To translate the raw acceleration outputs into executable motion commands by the agent, each value is combined with the previous velocity using a weighted sum. This approach ensures smooth transitions between steps. The resulting velocity updates are clipped to maintain valid bounds and are given by
\begin{align*}
\begin{array}{l}
v_t = \text{clip}\left(v_{t-1} + \beta_v \cdot a,\ -1,\ 1\right) \\
w_t = \text{clip}\left(\omega_{t-1} + \beta_\omega \cdot \alpha,\ -1,\ 1\right),
\end{array}
\end{align*}
where $v_t$ represents the velocity at time $t$, and $\omega_t$ the angular velocity at time $t$, and $\beta_v$, $\beta_\omega$ are scalar coefficients used to smooth the acceleration actions over several steps. 

In our experiments, both $\beta$ values were set to 0.5. The resulting velocities are then scaled by their respective maximum values, with $v$ bounded by $\pm0.5m/s$ in MiniWorld and $\pm1.0m/s$ in Gazebo with $\omega$ by $\pm \frac{\pi}{4} \text{ rad/s}$ in both scenarios. This configuration allows the agent to accelerate by a maximum of $\pm0.25 m/s$ in MiniWorld and $\pm0.5m/s$ in Gazebo and $\pm \frac{\pi}{8} \text{ rad/s}$ where $s$ relates to a step in MiniWorld and approximately one simulated second within Gazebo. 

As the agent interacts with the scenario using a continuous action space, Gazebo accommodates this by providing Twist commands consisting of angular and linear velocities. However, in the case of MiniWorld the provided action space uses discrete actions consisting of: $turn\_left$, $turn\_right$, $move\_forward$ and $move\_backward$. To update the position of the agent using a continuous action, the agents movement is parameterized by $v$ and $\omega$. 

Here, $v$ represents the length of the arc across a circle with a central angle of $\omega$ (in radians) between the initial and final position. With this we can then compute the radius of the circular arc as 
\[r={v \over \omega},\]
\noindent and the chord length $d$ as
\[d=2rsin(\omega/2).\]
The angle $\alpha$ between the chord and radius from the circle center to the agent's initial or final position as
\[\alpha={\pi \over 2} -{\omega \over 2}.\]
Translating this to the 2D grid within MiniWorld, the next position can be found given the initial position $x_0, y_0$ and orientation to the global reference frame $\theta_o$. This produces the next $x$ coordinate as
\begin{align*}
x &= x_0 + d \cos\left(\theta_0 + \frac{\omega}{2}\right) \\
  &= x_0 + r \left(\sin(\theta_0 + \omega) - \sin(\theta_0)\right),
\end{align*}
and similarly for the $y$ coordinate
\begin{align*}
y &= y_0 + d \sin\left(\theta_0 + \frac{\omega}{2}\right) \\
  &= y_0 - r \left(\cos(\theta_0 + \omega) - \cos(\theta_0)\right).
\end{align*}
The agents final orientation $\theta$ is then
\begin{align*}
\theta &= \theta_0 + \omega.
\end{align*}
This movement method for updating the coordinates $x,y$ and angle $\theta$ are then integrated within MiniWorld to replace the original discrete actions producing an agent that can be controlled using continuous actions resembling movements similar to that of robotic agents.

\subsection{Reinforcement Learning Policies and Training} \label{subsec:abstract_reinforcement_policies}

\textbf{Reinforcement Learning Policies}
To begin, the inputs for each modality are outlined below. Each modality additionally receives a 2-dimensional input vector representing the agent's current velocity, as well as a 36-dimensional LiDAR input vector. The LiDAR readings are averaged across discrete angular regions within \textit{Gazebo}, clipped to a maximum range of 5\,m, and normalized to the range $[0, 1]$.

\begin{enumerate}
    \item \textbf{Raw Image:} A $96 \times 96 \times 3$ RGB image is provided directly to the network.
    
    \item \textbf{YOLO:} To reduce complexity, only the $x$-coordinate of the detected object's centroid (with respect to the agent's egocentric view) is used. This value is encoded using sine and cosine functions to maintain continuity across image boundaries and to enable smooth transitions as the object moves across the field of view, resulting in a 2-dimensional input vector.
    
    \item \textbf{Position:} The agent's current $[x, y]$ position is normalized with respect to environment bounds and encoded using sine and cosine, yielding a final 4-dimensional input vector.
    
    \item \textbf{Grid Cell:} A 72-dimensional grid cell activation vector is concatenated with sine and cosine encodings of the agent's orientation, producing a 74-dimensional input vector.
\end{enumerate}

Using these modalities, the policies interact with the scenarios using and the Gymnasium reinforcement learning interface~\cite{towers2024gymnasiumstandardinterfacereinforcement} which is natively implemented within the MiniWorld simulator. In contrast, Robot control and perception are integrated through topics and services provided by the Robot Operating System (ROS)\cite{quigley2009ros} within Gazebo and operate in pseudo real time, to account for this with our experiments we found that a step size of $0.05sim\_seconds$ balances training performance and agent stability. 

In MiniWorld, due to the simplified nature of the simulation environment, the policy architecture is designed with multiple fully connected layers shared between the actor and critic networks. For simplicity, the actor outputs a two-dimensional action vector, while the critic outputs a single scalar value. When using vector-based observations, the policy consists of a 128-dimensional input layer followed by two hidden layers of 128 neurons each, producing a 64-dimensional feature representation. Gaussian Error Linear Unit (GeLU) activations are applied between the hidden layers.

In contrast, for image-based observations, the policy incorporates a convolutional encoder followed by a fully connected feature projection. The encoder comprises three convolutional layers with ReLU activations and increasing channel dimensions of 32, 64, and 64. These layers use kernel sizes of $8 \times 8$, $4 \times 4$, and $3 \times 3$, and corresponding strides of $4 \times 4$, $2 \times 2$, and $1 \times 1$, respectively. Given an input image of size $96 \times 96 \times 3$, the final flattened output of the convolutional stack is a 4096-dimensional feature vector, which is then passed through a fully connected layer with 256 neurons to produce the final policy features.

Within Gazebo, a similar model architecture is employed, with the primary difference being the replacement of the PPO core with a recurrent-PPO LSTM-based structure. Additionally, the actor and critic networks are implemented as separate models. The convolutional feature extractor is retained for image-based inputs, while the vector-based policy undergoes a reduction of one hidden layer and a doubling of the hidden dimension, resulting in a feature extractor composed of layers with 256, 256, and 128 units.

The extracted features—whether from the image or vector-based policy—are passed through a single-layer LSTM with a hidden dimension of 128. The LSTM output is then processed by a two-layer fully connected network with 128 and 64 neurons, respectively, and TanH activation functions between layers. The final output layer also uses a TanH activation, encouraging the network to produce outputs within the $[-1, 1]$ range, which aligns with the bounds of the action space and reward function. With the policies constructed the resulting model size given the varying input modalities is displayed within column 2 of Table \ref{table:model_sizes}.

\begin{table}[h]
\centering
\caption{Model sizes for the reinforcement learning policy, pre trained extractor and LSTM encoder used during training for the varying input modalities.}
\begin{tabular}{|ccccc|}
\hline
\multicolumn{5}{|c|}{\textbf{MINIWORLD}}                                                                                                                                                                                                                                                                                               \\ \hline
\multicolumn{1}{|c|}{\textbf{Input Type}} & \multicolumn{1}{c|}{\textbf{\begin{tabular}[c]{@{}c@{}}RL\\  Policy\end{tabular}}} & \multicolumn{1}{c|}{\textbf{\begin{tabular}[c]{@{}c@{}}Pre-Trained\\ Extractor\end{tabular}}} & \multicolumn{1}{c|}{\textbf{\begin{tabular}[c]{@{}c@{}}LSTM\\ Encoder\end{tabular}}} & \textbf{Total} \\ \hline
\multicolumn{1}{|c|}{Image}      & \multicolumn{1}{c|}{4.29}                                                          & \multicolumn{1}{c|}{-}                                                                        & \multicolumn{1}{c|}{-}                                                               & \textbf{4.29}  \\ \hline
\multicolumn{1}{|c|}{YOLO}                & \multicolumn{1}{c|}{0.16}                                                          & \multicolumn{1}{c|}{11.49}                                                                    & \multicolumn{1}{c|}{-}                                                               & \textbf{11.65} \\ \hline
\multicolumn{1}{|c|}{Position}            & \multicolumn{1}{c|}{0.16}                                                          & \multicolumn{1}{c|}{-}                                                                        & \multicolumn{1}{c|}{-}                                                               & -              \\ \hline
\multicolumn{1}{|c|}{\textbf{EDEN (OURS)}}           & \multicolumn{1}{c|}{0.2}                                                           & \multicolumn{1}{c|}{-}                                                                        & \multicolumn{1}{c|}{0.7}                                                             & \textbf{0.9}   \\ \hline
\multicolumn{5}{|c|}{\textbf{GAZEBO}}                                                                                                                                                                                                                                                                                                  \\ \hline
\multicolumn{1}{|c|}{\textbf{Input Type}} & \multicolumn{1}{c|}{\textbf{RL}}                                                   & \multicolumn{1}{c|}{\textbf{Extractor}}                                                       & \multicolumn{1}{c|}{\textbf{Predictor}}                                              & Total          \\ \hline
\multicolumn{1}{|c|}{Image}               & \multicolumn{1}{c|}{10.51}                                                         & \multicolumn{1}{c|}{-}                                                                        & \multicolumn{1}{c|}{-}                                                               & \textbf{10.51} \\ \hline
\multicolumn{1}{|c|}{YOLO}                & \multicolumn{1}{c|}{1.94}                                                          & \multicolumn{1}{c|}{10.13}                                                                    & \multicolumn{1}{c|}{-}                                                               & \textbf{12.07} \\ \hline
\multicolumn{1}{|c|}{Position}            & \multicolumn{1}{c|}{1.94}                                                          & \multicolumn{1}{c|}{-}                                                                        & \multicolumn{1}{c|}{-}                                                               & -              \\ \hline
\multicolumn{1}{|c|}{\textbf{EDEN (OURS)}}           & \multicolumn{1}{c|}{2.08}                                                          & \multicolumn{1}{c|}{83.21}                                                                    & \multicolumn{1}{c|}{1.77}                                                            & \textbf{87.06} \\ \hline
\end{tabular}
\label{table:model_sizes}
\end{table}

To provide an overview of the resulting model sizes, Column 2 of Table~\ref{table:model_sizes} reports the additional parameters introduced by the pretrained extractors used for object detection or feature extraction. In the MiniWorld environment, a finetuned YOLO model is employed to detect the blue box goal. In the Gazebo setup, a YoloV5n model is used for object detection, while the DINOv2 model serves as a general-purpose feature extractor within the \sys{} grid cell encoder. For the MiniWorld configuration, AprilTag detection is performed using a traditional algorithmic method; therefore, no additional parameters are reported for this component.

\textbf{Reward System and Training}
To complement the PPO algorithm, a reward system is defined consisting of three main components; per-step reward ($r_s$), upon collision ($r_c$), and reaching the goal ($r_g$). This reward structure incentivizes the agent to reach the goal efficiently while minimizing collisions. For efficient and reliable navigation, we employ a sparse reward setting with $r_g=(1-0.2\times({steps\over max\_steps}))$ where $max\_steps$ denotes the maximum number of steps before episode truncation and $steps$ refers to the number of steps taken in the current episode, $r_c=-0.1$, and $r_s=-0.01$. 

For stable learning and efficient use of available resources, the agent is trained using 10 parallel instantiations for MiniWorld and 6 parallel instantiations in Gazebo the generated plots use a statistic window size of 50 within Miniworld and 30 within Gazebo relating to $5\times$ the number of parallel instantiations. During training, the reinforcement learning policies were trained with a rollout size of 1024 steps and a batch size of 128 in MiniWorld, and a rollout size of 512 steps in Gazebo. To promote effective exploration and stable learning, the following hyperparameters were used: an entropy coefficient of 0.01, a discount factor $\gamma$ of 0.99, a GAE ($\lambda$) value of 0.95, actor and critic loss coefficients of 0.5, gradient norm clipping with a threshold of 0.5, and a clipping range of 0.2.

\subsection{Distance Normalized Steps} \label{subsec:abstract_norm_steps}
To outline the distance-normalized steps provided in Figure~\ref{fig:miniworld_gazebo_successrate}, let us take the example of the MiniWorld complex scenario where the agent's current position is denoted by $(x, y)$, and the positions of the three doors are given by:
\[
D_1 = (x_1, y_1), \quad D_2 = (x_2, y_2), \quad D_3 = (x_3, y_3)
\]
We define the Manhattan distance to each door as:
\[
d_i = |x - x_i| + |y - y_i|, \quad \text{for } i \in \{1, 2, 3\}
\]
he total Manhattan distance to all three doors is then:
\[
d_{\text{total}} = d_1 + d_2 + d_3
\]
Given that the agent takes $s$ steps to reach the goal, the distance-normalized steps are computed as:
\[
s_{\text{normalized}} = \frac{s}{d_{\text{total}}}
\]
 A similar procedure is repeated within Gazebo and the open room structures producing a metric that provides a measure of navigation efficiency relative to the overall structural layout of the scenario.

\subsection{Grid Cell Encoder and Training}\label{subsec:abstract_gc_encoder}
To train the grid cell encoder, a representative dataset was constructed for supervised sample generation. In \textit{MiniWorld}, the agent was trained across 20{,}000 trajectories of 400 steps each, while in \textit{Gazebo}, 1{,}280 trajectories of 1{,}000 steps were collected. Using these datasets, a decoder model was trained based on the following input modalities:

\begin{itemize}
    \item \textbf{MiniWorld AprilTags:} At each timestep, the current tag ID and the coordinates of its four corners are recorded. The corner coordinates are encoded using sine and cosine representations for rotational stability, resulting in a 16-dimensional vector. A total of 22 AprilTags are placed throughout the environment and represented using one-hot encoding, producing a final input vector of size 38.
    
    \item \textbf{Gazebo DINO Features:} For Gazebo, visual features are extracted using a pretrained DINOv2 feature extractor, resulting in a 384-dimensional feature vector per observation.
\end{itemize}

In both cases, the agent's velocity is appended to the visual features, resulting in a final input dimensionality of 40 for MiniWorld and 386 for Gazebo.

\textbf{Encoder Architecture:}  
In \textit{MiniWorld}, the encoder network begins with a two-layer feedforward network with 128 hidden units per layer, using the Exponential Linear Unit (ELU) activation function. This is followed by an LSTM with a hidden size of 72, corresponding to the grid cell encoding dimension. The final output stage consists of a three-layer feedforward network with dimensions 256, 256, and 74, again using ELU activations between layers. A final \texttt{Tanh} activation is applied to bound the outputs to the range $[-1, 1]$. The final output vector contains 72 grid cell activations and a 2-dimensional sine and cosine encoding of the agent’s orientation.

In \textit{Gazebo}, a similar architecture is adopted, with the following modifications: ELU activations are replaced with Gaussian Error Linear Units (GELU), and the input layers are expanded to dimensions 512 and 256. Layer normalization is applied after the initial layers for regularization. The output head also consists of 256, 256 layers with 20\% dropout applied between them, followed by a final 74-dimensional output layer. Regularization techniques are applied to the gazebo case due to the reduced amount of available samples.

\begin{figure}[h]
    \centering
    \includegraphics[width=\linewidth]{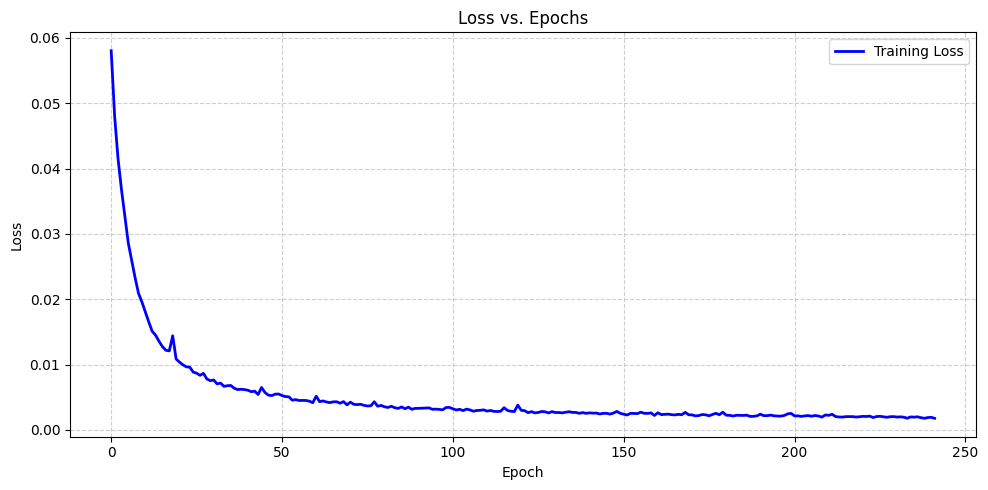}
    \caption{Training loss for encoding AprilTags within Miniworld into grid code}
    \label{fig:tag_loss}
\end{figure}

\textbf{Encoder Training and Results:}  
To ensure robustness and improve generalization, the encoder module is trained on sequences of varying lengths. Training is conducted for 250 epochs in \textit{MiniWorld} and 500 epochs in \textit{Gazebo}. To initialize the LSTM hidden state at the start of each batch, a weighted combination is used: 
\[
h_0 = 0.8 \times g_0 + 0.2 \times \theta_0,
\]
where $g_0$ denotes the initial grid cell activation, and $\theta_0$ represents the initial orientation relative to a global reference frame.

\begin{figure}[h]
    \centering
    \includegraphics[width=\linewidth]{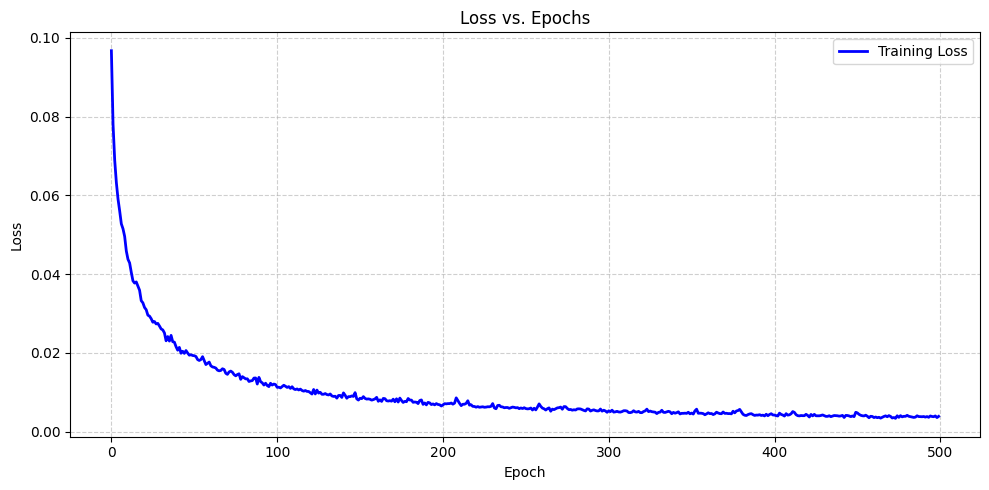}
    \caption{Training loss for encoding DINO features within Gazebo into grid cells}
    \label{fig:dino_loss}
\end{figure}

The models are optimized using a Mean Squared Error (MSE) loss to predict the target grid cell encodings. Training results are shown in Figure~\ref{fig:tag_loss} for the MiniWorld AprilTag-based decoder, and in Figure~\ref{fig:dino_loss} for the Gazebo DINO-based decoder showing smooth convergence to predict grid cell activations.

\vspace{1em} 

As shown and discussed in Figure~\ref{fig:decoder_results}, the trained grid cell encoder is evaluated by deploying a policy-driven and random acting agent within the scenarios. Multiple trajectories are collected, and the movable area is divided into 100 equal regions. Histogram binning is then applied to generate the final heatmap, which serves as a qualitative tool to confirm the successful training of the encoder.

\end{document}